\documentclass[journal,twocolumn,draftclsnofoot]{IEEEtran}

\usepackage{amsmath,amssymb,amsfonts}
\usepackage{algorithmic}
\usepackage{graphicx}
\usepackage{textcomp}
\usepackage{xcolor}
\usepackage{mathtools}
\usepackage{multirow}
\usepackage{subfigure}

\ifCLASSINFOpdf
\else
\fi

\usepackage{bbm, dsfont}

\begin{document}
%
\title{Bad and good errors: \\value-weighted skill scores in deep ensemble learning}
%
%
%

\author{Sabrina~Guastavino,
        Michele~Piana,
        and~Federico~Benvenuto%
\thanks{Sabrina Guastavino, Michele Piana and Federico Benvenuto are with the Dipartimento di Matematica, Università di Genova (e-mail: guastavino@dima.unige.it; piana@dima.unige.it; benvenuto@dima.unige.it ).}
\thanks{Manuscript received ..; revised ...}}

%
%

\markboth{Journal of \LaTeX\ Class Files,~Vol.~14, No.~8, August~2015}%
{Shell \MakeLowercase{\textit{et al.}}: Bare Demo of IEEEtran.cls for IEEE Journals}
%



\maketitle

\begin{abstract}
In this paper we propose a novel approach to realize forecast verification. Specifically, we introduce a strategy for assessing the severity of forecast errors based on the evidence
that, on the one hand, a false alarm just anticipating an occurring event is better than one in the middle of consecutive non-occurring events, and that, on the other hand, a miss of an isolated event has a worse impact than a miss of a single event, which is part of several consecutive occurrences.
Relying on this idea, we introduce
a novel definition of confusion matrix and skill scores giving greater importance to the value of the prediction rather than to its quality. 
Then, we introduce a deep ensemble learning procedure for binary classification, in which the probabilistic outcomes of a neural network are clustered via optimization of these value-weighted skill scores. 
We finally show the performances of this approach in the case of three applications concerned with pollution, space weather and stock prize forecasting.
\end{abstract}


%
\IEEEpeerreviewmaketitle

\section{Introduction}
\IEEEPARstart{P}{redicting} events over time applies to a number of fields, ranging from weather \cite{gneiting2005weather} and space weather forecasting \cite{camporeale2018machine}, through environment \cite{kurt2008online} to stock market forecasting \cite{deepak2017machine}. In all these frameworks, the goodness of prediction is usually defined in terms of the correspondence between forecasts and observations and is known in the literature as the {\it forecast quality} \cite{1993WtFor...8..281M}. For binary predictions, forecast quality is typically measured by using skill scores based on a confusion matrix whose entries count the number of false and true negatives and of false and true positives. Specifically, these skill scores rely on simple arithmetic formulas that compute in different ways the imbalance of the diagonal entries (representing the correct prediction) with respect to the off-diagonal entries (representing the incorrect prediction). Typical skill scores for forecast quality are the True Skill Statistic (TSS) \cite{allouche2006assessing}, the Heidke Skill Score (HSS) \cite{hyvarinen2014probabilistic} and the Critical Success Index (CSI) \cite{schaefer1990critical}. However, there is a different perspective to evaluate forecast, i.e., in terms of its usefulness to support the user while making a decision.
This type of goodness is known in the literature as the {\it forecast value} \cite{mylne2002decision} and two examples of scores for the quantitative evaluation of this prediction goodness are the Cost Value Score and the Relative Value Score introduced in \cite{R-98}.
Moreover, the field of cost-sensitive learning is devoted to take the misclassification costs (and possibly other types of cost) into consideration \cite{elkan2001foundations}.
These skill scores give an extrinsic value of the prediction as they are based on a cost-benefits ratio belonging to the sphere of decision making processes outer of the prediction itself. To compute such skill scores is necessary to quantify the cost, the loss and the benefits of taking (or not) an action.

In this paper we focus on binary predictions performed over time and we propose a novel approach to evaluate the severity of prediction errors by considering that a false alarm predicting that an event will occur just before its actual occurrence, anyway eases the right decision from the user, while a delayed alarm is of little use. In this way, we classify errors on the basis of their importance and we get a novel notion of confusion matrix and related skill scores. In this new framework, the severity of errors depends on their impact on the decision making process and therefore we refer to these novel confusion matrix and skill scores as {\em{value-weighted}}.
Operationally, we propose a general strategy for selecting optimal predictions among the ones provided by the training of a deep neural network. This strategy is inspired by ensemble learning, relies on the novel skill scores and implements the following steps:
\begin{itemize}
    \item A deep neural network is trained on a historical data set, thus providing a set of probabilistic forecasts (indexed over the epochs of the training process) almost equivalent in terms of forecast quality but different from each other with respect to forecast value;
    \item The clustering of the probabilistic outcomes of the neural network is realized by maximizing a given skill score on the training set, so that the probabilistic forecasts    indexed over the epochs are transformed into binary predictions;
    \item These binary predictions are exploited within an ensemble learning framework and the approach is applied on the validation set.
\end{itemize}

This ensemble network forecasting, when optimized with the value-weighted skill scores, yields better results in terms of both forecast quality and value, than the ones provided by the optimization of classical quality-based skill scores. We showed this better effectiveness in the case of three applications. 
We considered the problem of forecasting the air quality in highly polluted urban environments, with particular reference to the prediction of concentration of fine particles with diameters of $2.5$ $\mu$m and smaller (PM2.5) \cite{perez2000prediction,zamani2019pm2,pak2020deep}. The second problem was concerned with solar flare forecasting, i.e. the prediction of those explosive solar events that trigger most space weather phenomena \cite{bobra2015solar,benvenuto2018hybrid,nishizuka2018deep}. 
Eventually, the third application focuses on stock prize forecasting and, in particular, on the prediction of ``down" movements in the market \cite{bustos2020stock,liu2020recurrent,shi2020stock}.

The plan of the paper is as follows.
In Section \ref{sec:II} we define the classical skill scores and we show that binary predictions of the same quality can have completely different forecast values.
In Section \ref{sec:III} we introduce a value-weighted confusion matrix and the corresponding value-weighted skill scores to quantify the forecast value of a binary prediction. In Section \ref{sec:IV} we show how this value-weighted approach works to assess the performances of standard machine learning. Section \ref{sec:V} introduces the ensemble learning process and describes its application to pollution, space weather and stock prize forecasting. Our conclusions are offered in Section \ref{sec:VI}.

\section{Confusion matrix and skill scores}
\label{sec:II}
The results of a binary classifier are usually evaluated by computing the
confusion matrix, also known as contingency table. 
Let $\mathbb{M}_{2,2}(\mathbb{N})$ be the set of $2$-dimensional matrices with integers elements. Let $\mathbf{y}\in\{0,1\}^n$ be a binary vector representing the true label vector and let $\mathbf{p}\in\{0,1\}^n$ be a binary prediction. Then the confusion matrix $\mathbf{C}\in\mathbb{M}_{2,2}(\mathbb{N})$ is defined as: 
\begin{itemize}
    \item $\mathbf{C}_{1,1}=\#\{i\in\{1,\dots,n\} : y_i=1, p_i=1 \}$. 
    \item $\mathbf{C}_{2,2}=\#\{i\in\{1,\dots,n\} : y_i=0, p_i=0 \}$. 
    \item $\mathbf{C}_{1,2}=\#\{i\in\{1,\dots,n\} : y_i=0, p_i=1 \}$. 
    \item $\mathbf{C}_{2,1}=\#\{i\in\{1,\dots,n\} : y_i=1, p_i=0 \}$. 
\end{itemize}
This definition implies that $\mathbf{C}_{1,1}$ computes the True Positives (TPs), $\mathbf{C}_{2,2}$ computes the True Negatives (TNs), $\mathbf{C}_{1,2}$ computes the False Positives (FPs) and $\mathbf{C}_{2,1}$ computes the False Negatives (FNs). From this confusion matrix several skill scores can be computed in order to evaluate the binary classifier performances. Given a confusion matrix $\mathbf{C}$, we denote with $S:\mathbb{M}_{2,2}(\mathbb{N})\to\mathbb{R}$ a skill score defined on the matrix $\mathbf{C}$. Four frequently used skill scores are: 
\begin{itemize}
    \item Accuracy (ACC) \cite{sokolova2006beyond}: 
    \begin{equation}
        ACC(\mathbf{C})=\frac{\mathbf{C}_{1,1}+\mathbf{C}_{2,2}}{\mathbf{C}_{1,1}+\mathbf{C}_{1,2}+\mathbf{C}_{2,1}+\mathbf{C}_{2,2}}~,
    \end{equation}
    i.e., the ratio between the number of correct predictions over the total number of
predictions. $ACC(\mathbf{C})\in [0,1]$
and the optimal value is $1$.
\item True Skill Statistic (TSS) \cite{hanssen1965relationship}:
\begin{equation}
    TSS(\mathbf{C})=\frac{\mathbf{C}_{1,1}}{\mathbf{C}_{1,1}+\mathbf{C}_{2,1}}-\frac{\mathbf{C}_{1,2}}{\mathbf{C}_{1,2}+\mathbf{C}_{2,2}}~,
\end{equation}
i.e., the balance between the true positive rate (or probability of detection) and the false alarm rate. $TSS(\mathbf{C})\in [-1,1]$ and it is optimal when
it is equal to 1. A negative value means that forecasting behaves in a wrong
way i.e. it mixes the role of the positive events with the role of the negative ones.
\item Heidke Skill Score (HSS) \cite{Heidke1926}:
\begin{equation}
    HSS(\mathbf{C})=\\
    \dfrac{2(\mathbf{C}_{1,1}\mathbf{C}_{2,2}-\mathbf{C}_{2,1}\mathbf{C}_{1,2})}{\mathbf{T}_1+\mathbf{T}_2}~,
\end{equation}
where $\mathbf{T}_1 := (\mathbf{C}_{1,1}+\mathbf{C}_{2,1})(\mathbf{C}_{2,1}+\mathbf{C}_{2,2})$ and $\mathbf{T}_2 := (\mathbf{C}_{1,1}+\mathbf{C}_{1,2})(\mathbf{C}_{1,2}+\mathbf{C}_{2,2})$,
i.e., a measure of the improvement of forecast over random forecast. $HSS(\mathbf{C})\in (-\infty,1]$. The optimal value is equal to $1$, a
negative value meaning that forecast is worse than random forecast and the $0$ value meaning that the forecast has the same skill of random forecast.
\item Critical Success Index (CSI) \cite{schaefer1990critical}:
\begin{equation}
    CSI(\mathbf{C})=\frac{\mathbf{C}_{1,1}}{\mathbf{C}_{1,1}+\mathbf{C}_{1,2}+\mathbf{C}_{2,1}},
\end{equation}
i.e., the ratio between the number of correct event forecast and the number of
events which occurred plus the number of
false alarms. $CSI(\mathbf{C})\in [0,1]$ and the optimal value is equal to $1$.
\end{itemize} 

Given a binary-vector $\mathbf{y}\in\{0,1\}^{n}$ encoding the binary outcome of an empirical observation, we can define a function $F_{\mathbf{y}}$ such as
\begin{equation}
\begin{split}
\label{map F_y}
F_{\mathbf{y}}: &  \{0,1\}^n \to \mathbb{M}_{2,2}(\mathbb{N}) \\ 
& \mathbf{p} \mapsto  C,
\end{split}
\end{equation}
which maps a binary prediction $\mathbf{p}\in\{0,1\}^{n}$ onto the confusion matrix obtained by comparing $\mathbf{y}$ and $\mathbf{p}$. The function $F_{\mathbf{y}}$ is clearly not injective.  
\begin{figure*}
    \centering
    \includegraphics[width=0.7\textwidth]{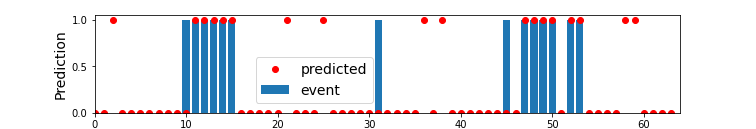}\\
    \includegraphics[width=0.7\textwidth]{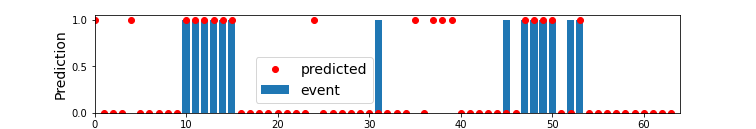}\\
    \includegraphics[width=0.7\textwidth]{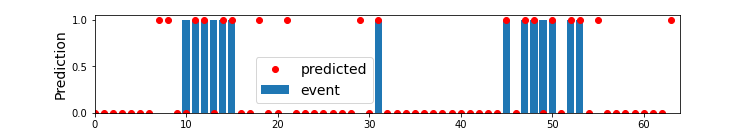}\\
     \includegraphics[width=0.7\textwidth]{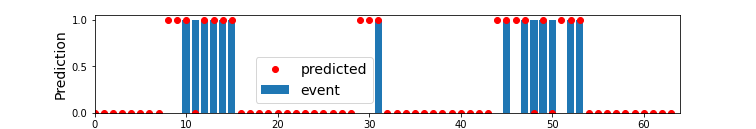}
    \caption{Four different binary predictions with the same confusion matrix characterized by $\mathbf{C}_{1,1}=11$ (TPs), $\mathbf{C}_{1,2}=7$ (FPs), $\mathbf{C}_{2,1}=3$ (FNs) and $\mathbf{C}_{2,2}=43$ (TNs).}
    \label{fig:toy_ex}
\end{figure*}
Figure \ref{fig:toy_ex} illustrates an example in which, given a binary observation $\mathbf{y}$, four different binary predictions lead to the same confusion matrix. In the example, ${\bf{y}}$ has $64$ components such that $14$ components are equal to $1$ and $50$ components are equal to zero. The four predictions $\mathbf{p}^{(1)},\mathbf{p}^{(2)}, \mathbf{p}^{(3)}$ and $\mathbf{p}^{(4)}$ in the four panels of the figure provide the same confusion matrix with entries TP=$11$, FN=$3$, FP=$7$ and TN=$43$. From a forecasting quality viewpoint, the four predictions are the same, since they lead to the same confusion matrix and, accordingly, to the same skill scores. However, from a forecasting value viewpoint, the four predictions are different. More specifically, prediction $\mathbf{p}^{(4)}$, for which the corresponding FPs closely anticipate the observed outcomes equal to $1$ should be preferred, in value terms, than the other predictions that sound alarms after the occurrence of the events. 

We now introduce a novel definition of confusion matrix, which is able to distinguish between these ambiguous configurations and therefore to locally restore the injectivity of $F_{\mathbf{y}}$.

\section{Value-weighted Confusion matrix and skill scores}
\label{sec:III}
In order to define the value-weighted confusion matrix, we first introduce the error functions 
$\varepsilon_{1,2}:\{0,1\}\times \{0,1\}\to\mathbb{R}_+$ and $\varepsilon_{2,1}:\{0,1\}\times \{0,1\}\to\mathbb{R}_+$ such that, given the
component $y_i$ of a binary observed outcome ${\bf{y}}$ and the component $p_i$ of a binary prediction ${\bf{p}}$, $\varepsilon_{1,2}(y_i,p_i)$ and $\varepsilon_{2,1}(y_i,p_i)$ measure the error of the incorrect prediction when either an FP or an FN occurs. These error functions allow the generalization of the confusion matrix concept as follows. 
We denote with $\mathbbm{1}_{\{a\}}$ a number equal to $1$ when condition $a$ is satisfied and $0$ otherwise. 
Then the weighted confusion matrix $\tilde{\mathbf{C}}\in\mathbb{M}_{2,2}(\mathbb{R_+})$ is defined as
\begin{equation}
    \tilde{\mathbf{C}}_{1,1}=\sum_{i=1}^n \mathbbm{1}_{\{y_i=1,p_i=1\}}, \quad \tilde{\mathbf{C}}_{2,2}=\sum_{i=1}^n \mathbbm{1}_{\{y_i=0,p_i=0\}}, 
\end{equation}
\begin{equation}
    \tilde{\mathbf{C}}_{1,2}=\sum_{i=1}^n \varepsilon_{1,2}(y_i,p_i), \quad   \tilde{\mathbf{C}}_{2,1}=\sum_{i=1}^n \varepsilon_{2,1}(y_i,p_i) ~.
\end{equation}
On the one hand, quality-based forecasting assumes
\begin{eqnarray}\label{quality-based}
\varepsilon_{1,2}(y_i,p_i):=\mathbbm{1}_{\{y_i=0,p_i=1\}}, ~
\varepsilon_{2,1}(y_i,p_i):=\mathbbm{1}_{\{y_i=1,p_i=0\}}.
\end{eqnarray} 
On the other hand, in the case of value-weighted forecasting we choose
\begin{equation}\label{eq:psi-1}
\varepsilon_{1,2}(y_i,p_i)=\psi(y_i,p_i) \mathbbm{1}_{\{y_i=0,p_i=1\}}
\end{equation}
with
\begin{equation}\label{eq:psi-2}
\psi(y_i,p_i)=
\begin{cases}
1-\displaystyle\max_{1\le k\le K}\left(\dfrac{y_{i+k}}{k+1}\right), & \text{ if }  1\in\{y_{i+t}\}_{t=-K}^{K}\\
2, 
& \mbox{ otherwise }~,
\end{cases}
\end{equation}
and
\begin{equation}\label{eq:phi-1}
\varepsilon_{2,1}(y_i,p_i)=\phi(y_i,p_i) \mathbbm{1}_{\{y_i=1,p_i=0\}}
\end{equation}
with
\begin{equation}\label{eq:phi-2}
\phi(y_i,p_i)=
\begin{cases}
1-\displaystyle\max_{1\le k\le K}\left(\frac{p_{i-k}}{k+1}\right), & \text{if} 1\in\{p_{i+t}\}_{t=-K}^{K}\\
2, 
& \mbox{ otherwise }~.
\end{cases}
\end{equation}
In equations (\ref{eq:psi-2}) and (\ref{eq:phi-2}), $K$ is a fixed positive integer number. Analogously to the case of the standard confusion matrix in Section \ref{sec:II}, we have that $\tilde{\mathbf{C}}_{1,1}$, $\tilde{\mathbf{C}}_{2,2}$, $\tilde{\mathbf{C}}_{1,2}$ and $\tilde{\mathbf{C}}_{2,1}$ compute, respectively the numbers of value-weighted true positives (wTPs), value-weighted true Negatives (wTNs), value-weighted false positives (wFPs), and value-weighted false negatives (wFNs).

We remark that $\tilde{\mathbf{C}}_{1,1},\tilde{\mathbf{C}}_{2,2} \in\mathbb{N}$ and that they have the same definition of the ones in the classical confusion matrix whereas $\tilde{\mathbf{C}}_{1,2},\tilde{\mathbf{C}}_{2,1}\in\mathbb{R}_+$ and can be seen as weighted version of $\mathbf{C}_{1,2}$ and $\mathbf{C}_{2,1}$ with weighting functions $\psi(y_i,p_i)$ and $\phi(y_i,p_i)$, respectively. 
In order to illustrate how these weighting functions work while computing the corresponding prediction error, we introduce the window
\begin{equation}\label{eq:window}
\mathcal{I}_{i,K}=\{i-K,\dots,i,\dots,i+K\}
\end{equation}
centered in the index $i$, with size $K$. Then, for the $i$-th sample we have two possible cases:
\begin{itemize}
\item Case of a false positive ($y_i=0$, $p_i=1$): $\psi(y_i,p_i)$ depends on the sequence $\{y_j\}_{j\in\mathcal{I}_{i,K}}$. In fact: 
\begin{itemize}
    \item If no event occurs in the window, i.e. $y_j=0$ for each $j \in \mathcal{I}_{i,K}$ then
    \begin{equation}
        \psi(y_i,p_i)=2.
    \end{equation}
\item If at least one event occurs in the window, i.e. $1\in\{y_j\}_{j\in\mathcal{I}_{i,K}}$, then 
\begin{equation}\label{psi_alarms}
\psi(y_i,p_i)=1-\max_{1\le k\le K}\left(\frac{1}{k+1}y_{i+k}\right)~.
\end{equation}
Therefore, from equation \eqref{psi_alarms} we can distinguish two situations:
\begin{itemize}
\item[1.] If the event occurs before time $i$ and there is no event in the next times, i.e. $y_j=0$ for $j\in\mathcal{I}_{i,K}$ and $j\ge i+1$, then
\begin{equation}
    \psi(y_i,p_i)=1,
\end{equation}
since in equation \eqref{psi_alarms} $\max_{1\le k\le K}(\frac{1}{k+1}y_{i+k})=0$;
\item[2.] If at least one event occurs after time $i$ then 
\begin{equation}
    \frac{1}{2}\le\psi(y_i,p_i)<1
\end{equation}
and the weighting function decreases according to the inverse of the distance of the next event occurrence, e.g. if $y_{i+1}=1$ then $\psi(y_i,p_i)=\frac{1}{2}$.
\end{itemize}
\end{itemize}
\end{itemize}
\begin{itemize}
\item Case of a false negative ($y_i=1$, $p_i=0$): $\phi(y_i,p_i)$ depends on the sequence $\{p_j\}_{j\in\mathcal{I}_{i,K}}$. In fact:
\begin{itemize}
    \item If no predicted alarm is in the window, i.e. $p_j=0$ for each $j\in\mathcal{I}_{i,K}$ then
    \begin{equation}
        \phi(y_i,p_i)=2.
    \end{equation}
\item If there is at least one predicted alarm in the window, i.e. $1\in\{p_j\}_{j\in\mathcal{I}_{i,K}}$, then 
\begin{equation}\label{eq:phi_alarms}
\phi(y_i,p_i)=1-\max_{1\le k\le K}\left(\frac{1}{k+1}p_{i-k}\right)~.
\end{equation}
In particular from equation \eqref{eq:phi_alarms} we can distinguish two situations:
\begin{itemize}
\item[1.] If the predicted alarm is after time $i$ and there are no predicted alarms in the previous times, i.e. $p_j=0$ for $j\in\mathcal{I}_{i,K}$ and $j\le i-1$, then
\begin{equation}
    \phi(y_i,p_i)=1,
\end{equation}
since $\max_{1\le k\le K}(\frac{1}{k+1}p_{i-k})=0$.
\item[2.] If there is at least one predicted alarm before time $i$, then 
\begin{equation}
    \frac{1}{2}\le\phi(y_i,p_i)<1
\end{equation}
and the weighting function decreases according to the inverse of the distance of the previous predicted alarm, e.g. if $p_{i-1}=1$ then $\phi(y_i,p_i)=\frac{1}{2}$.
\end{itemize}
\end{itemize}
\end{itemize}

We applied the value-weighted confusion matrix 
$\tilde{\mathbf{C}}$ relying on this error function on the motivating example in Figure 
\ref{fig:toy_ex}. The results in Table \ref{tab:toy_ex} inspires the following comments.

First, differently than what happens with the classical quality-based confusion matrix ${\bf{C}}$, the off-diagonal terms of $\tilde{\mathbf{C}}$ depend on the prediction vector. 

Second, the new confusion matrix 
gives a clearer idea on how the incorrect predictions are distributed while rolling them along the sample index. 
On the one hand, in the value-weighted approach, wFPs associated to $\bf{p}^{(1)}$ notably increase with respect to the quality-based FPs, coherently to the fact that this prediction sounds alarms far from the actual event occurrence. On the other hand, wFPs associated to prediction $\bf{p}^{(4)}$ significantly decrease, coherently to the fact that, in this case, many incorrectly predicted alarms anticipate the event occurrence. Similar considerations can be repeated for what concerns the three original samples incorrectly predicted as $0$: we notice that, in prediction $\mathbf{p}^{(4)}$, the number of wFNs is small, which means that the three missed events have been predicted in advance.
\begin{table}[h!]
\centering
\caption{Comparison between the value-weighted and quality-based confusion matrix and corresponding TSS for the motivating examples in Figure \ref{fig:toy_ex}.}\label{tab:toy_ex}
\resizebox{0.5\textwidth}{!}{  
\begin{tabular}{c | c c c | c c c}
Prediction & \multicolumn{3}{c|}{Value-weighted}  &  \multicolumn{3}{c} {} \\ 
& wFP & wFN & wTSS & FP & FN & TSS\\ \hline
$\mathbf{p}^{(1)}$ &  $14$ &  $4$ &  $0.4877$ & $7$ &  $3$ & $0.6457$\\ \hline 
$\mathbf{p}^{(2)}$ &  $14$ &  $3.67$ & $0.5044$ &   $7$ &  $3$ & $0.6457$\\  \hline 
$\mathbf{p}^{(3)}$ &  $8.08$ &  $1.67$ & $0.7102$ & $7$ &  $3$ & $0.6457$ \\ \hline 
$\mathbf{p}^{(4)}$ &  $3.83$ &  $1.5$ & $0.7981$ & $7$ &  $3$ & $0.6457$\\  
\hline
\end{tabular}
}
\end{table}
   
\section{Value-weighted skill scores in action}
\label{sec:IV}
\begin{figure*}
    \centering
     \subfigure[{PM2.5 concentration on period from 1/7/2013 00:00 to 1/15/2013 07:00.}]{
    \includegraphics[width=0.99\textwidth]{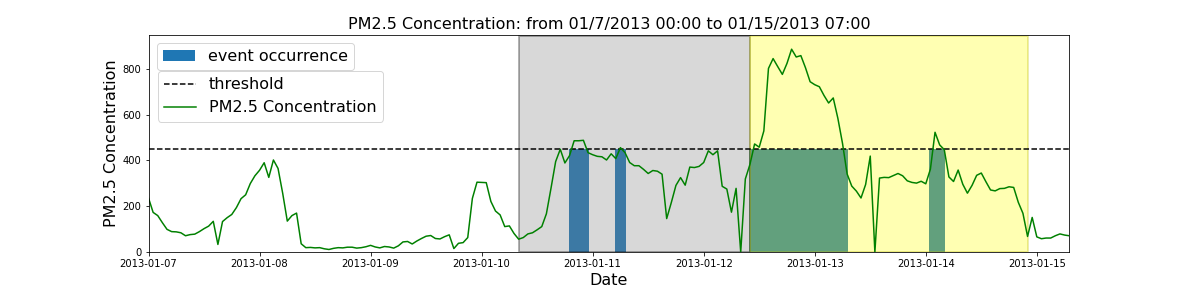}}\\
    \subfigure[{LR: prediction in grey box}]{\includegraphics[width=0.32\textwidth]{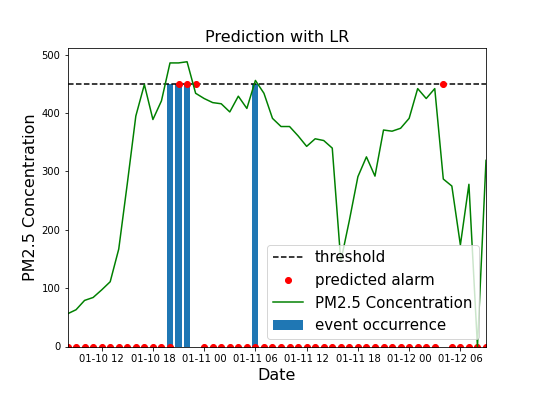}}
     \subfigure[{SVM: prediction in grey box }]{\includegraphics[width=0.32\textwidth]{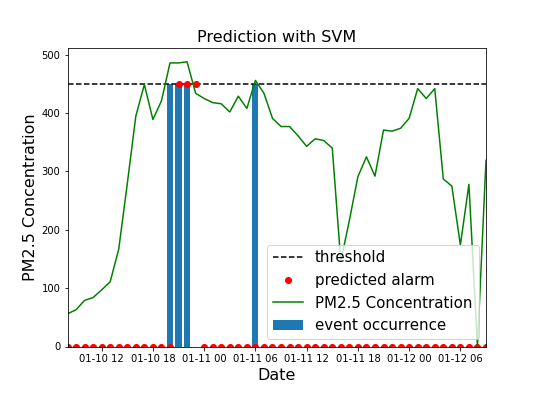}}
        \subfigure[{NN: prediction in grey box }]{ \includegraphics[width=0.32\textwidth]{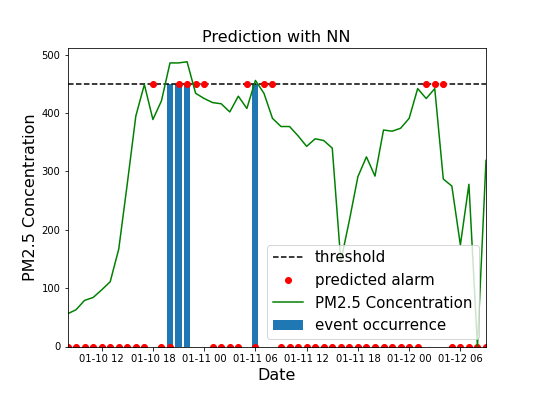}}\\
         \subfigure[{LR: prediction in yellow box}]{\includegraphics[width=0.32\textwidth]{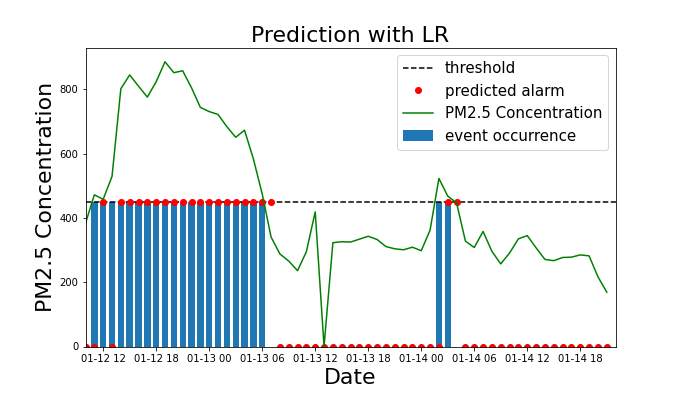}}
      \subfigure[{SVM: prediction in yellow box }]{\includegraphics[width=0.32\textwidth]{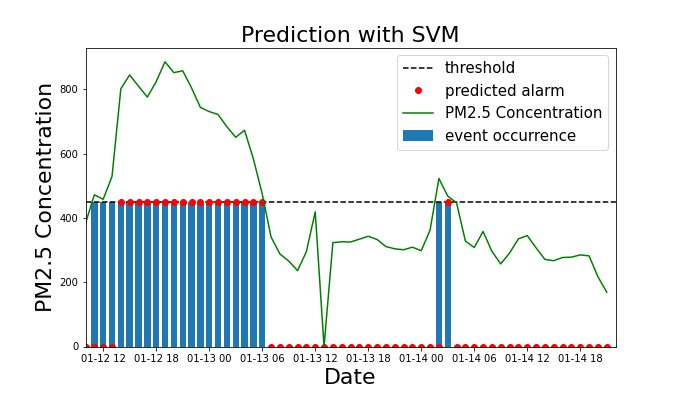}}
      \subfigure[{NN: prediction in yellow box}]{\includegraphics[width=0.32\textwidth]{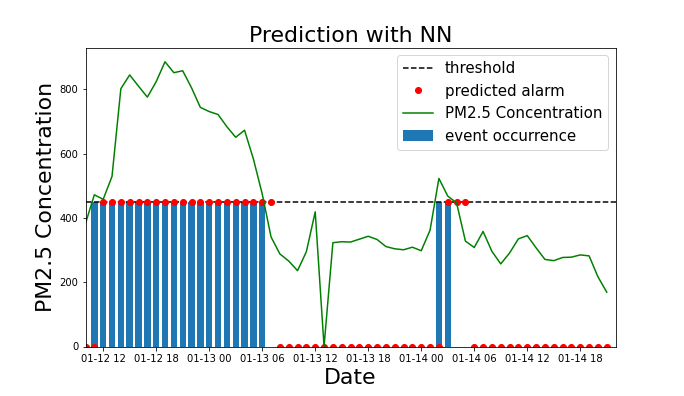}}
    \caption{Top panel: PM2.5 concentration distributed on the period from 1/7/2013 at 00:00 through 1/15/2013 at 07:00.
    Middle panels (from left to right): with reference to the grey box in the top panel, predictions enrolled along time from 1/10/2013 at 08:00 through 1/12/2013 at 09:00 provided by LR, SVM and NN, respectively. Bottom panels (from left to right): with reference to the yellow box in the top panel, predictions enrolled along time from 1/12/2013 at 10:00 through 1/14/2013 at 22:00 provided by LR, SVM and NN, respectively.}
    \label{fig:poll_evaluate_pred}
\end{figure*}

In order to test the behavior of this value-weighted approach in the case of experimental data associated to forecasting problems, we considered an example based on a data set from the University of California at Irvine (UCI), available at https://archive.ics.uci.edu/ml/datasets/Beijing+PM2.5+Data via the data interface released by the US embassy in Beijing \cite{Liang2015AssessingBP}. In detail, this archive includes:
\begin{itemize}
    \item Hourly weather information (dew point, temperature, pressure, wind direction and speed, cumulative number of snowing and raining hours).
    \item Hourly concentration of PM2.5.
\end{itemize}
The forecasting problem we considered is the one to predict whether PM2.5 concentration at time $T+1$ will exceed a fixed threshold associated to a condition of severely polluted air, given weather conditions and PM2.5 concentration at time $T$. The machine learning approaches used to address this forecasting problem were three standard supervised algorithms: Logistic Regression (LR) \cite{hosmer2013applied}, Support Vector Machine (SVM) \cite{cortes1995support} and a standard Neural Network (NN) \cite{Bishop1995Neural}. We trained and validated the three approaches using the dataset in the time range between 01/01/2010 and 12/28/2011 so that the training and validation sets had 17424 samples with just 122 samples labelled with 1 (corresponding to over-threshold pollution). 
Figure \ref{fig:poll_evaluate_pred} shows the forecasting provided by the three algorithms in the case of a test set in the archive corresponding to the time range between 1/7/2013 at 00:00 UT and 1/15/2013 at 07:00 UT. 
Table \ref{tab:pollution-evaluate-pred} allows a quantitative assessment of these performances by means of both the standard quality-based confusion matrix and corresponding TSS, and the confusion matrix and TSS provided by the value-weighted approach. 
We focused on TSS because we considered a significantly imbalanced training set and in this case this score is more appropriate \cite{bloomfield2012toward}.
A comparison between figure \ref{fig:poll_evaluate_pred} and table \ref{tab:pollution-evaluate-pred} shows how our value-weighted approach works. Indeed, looking at the table:
\begin{itemize}
    \item NN has significantly more FPs than the other two methods and slightly less FNs. As a consequence, its TSS is smaller than the one associated to LR (which produces a smaller number of FPs).
    \item NN has the highest wTSS, as a consequence of the smallest number of wFNs among the three methods.
\end{itemize}
Coherently with these values, Figure \ref{fig:poll_evaluate_pred} shows that NN sounds either timely alarms or alarms in advance with respect to the actual event occurrence (this is particularly true in the case of the window highlighted by the grey box). Further, it provides more FPs, but these are either sounded in a close neighborhood of the actual event occurrence or corresponds to a high PM2.5 concentration level.

\begin{table*}[ht]
		\centering
		\caption{Results provided by LR, SVM and NN in the case of the pollution forecasting experiment, when the test set is the UCI database in the time range from 1/7/2013 at 00:00 through 1/15/2013 at 07:00. 
}
\label{tab:pollution-evaluate-pred}
\resizebox{0.7\textwidth}{!}{
\begin{tabular}{l l l l l l l}
& \multicolumn{6}{c}{Method}\\
\cline{2-7}
 & \multicolumn{2}{c}{LR} &  \multicolumn{2}{c}{ SVM }&  \multicolumn{2}{c}{ NN }\\ 
Confusion matrix
& TP = 21 & FP = 4 & TP = 20  & FP = 1 & TP = 22 & FP = 12 \\ 
 & FN = 5  & TN = 170 & FN = 6  & TN = 173  & FN = 4 & TN = 162 \\ \hline 
TSS & \multicolumn{2}{c}{\textbf{0.7847}} &  \multicolumn{2}{c}{0.7635} & \multicolumn{2}{c}{0.7772}\\ 
wFN & \multicolumn{2}{c}{5.5} & \multicolumn{2}{c}{7} & \multicolumn{2}{c}{3.17}\\
wFP & \multicolumn{2}{c}{5} & \multicolumn{2}{c}{1} & \multicolumn{2}{c}{14.17}\\
wTSS & \multicolumn{2}{c}{0.7639} & \multicolumn{2}{c}{0.7350} & \multicolumn{2}{c}{\textbf{0.7938}}\\
\hline
\end{tabular}
}
\end{table*}

\section{Deep ensemble classifiers and applications}
\label{sec:V}
In deep learning, automatic classifiers can be constructed by applying thresholding procedures to neural networks (NNs) with probability outcomes. In this approach, a NN can be formally interpreted as the map
\begin{equation}\label{eq:NN}
    \theta(\mathbb{V},\cdot):\mathbb{R}^p\to [0,1],
\end{equation}
where $\mathbb{V}$ represents the space of weights and the ensemble learning process implements the following steps:
\begin{enumerate}
    \item Train $\theta(\mathbb{V},\cdot)$ on the training set $\{(\mathbf{x}_i,y_i)\}_{i=1}^n$ using an iterative optimization scheme that stops after $N$ epochs. 
    \item For each epoch $j$ and given $\mathbf{X}=(\mathbf{x}_1,\ldots,\mathbf{x}_n)$, choose the classification threshold as the real number that maximizes a specific skill score $S$. Therefore, if 
    \begin{equation}
        \hat{\mathbf{y}}_{\mathbf{X}}^{\mathbf{w}_j}:=(\theta(\mathbf{w}_j,\mathbf{x}_1),\dots, \theta(\mathbf{w}_j,\mathbf{x}_n))^T~,
    \end{equation}
    then the optimum threshold is the solution of the optimization problem
    \begin{equation}\label{opt tau at j-th epoch}
        \tau_j^* = \arg\max_{\tau\in [a,b]} S(F_{\mathbf{y}}(I_{\tau}(\hat{\mathbf{y}}_{\mathbf{X}}^{\mathbf{w}_j}))),
    \end{equation}
    where $[a,b]$ is a suitable interval with $0\le a<b\le 1$, $I_{\tau}:\mathbb{R}^n\to \{0,1\}^n$ is the indicator function
    $I_{\tau}(\hat{\mathbf{y}})=(\mathbbm{1}_{\{\hat{y}_i > \tau\}},\dots, \mathbbm{1}_{\{\hat{y}_n > \tau\}})^T$, $F_{\mathbf{y}}$, defined as in \eqref{map F_y}, maps the binary prediction to the associated confusion matrix computed with respect to the true label vector $\mathbf{y}$, and $S$ is the skill score computed on the confusion matrix, as defined in section \ref{sec:II}.
    \item Consider a validation set $\{(\tilde{\mathbf{x}}_i,\tilde{y}_i)\}_{i=1}^m$, and the matrix $\tilde{\mathbf{X}}$ such that $\tilde{\mathbf{X}}=(\tilde{\mathbf{x}}_1,\ldots,\tilde{\mathbf{x}}_m)$.
       For each epoch $j$, compute 
 \begin{equation}
        \hat{\mathbf{y}}_{\tilde{\mathbf{X}}}^{\mathbf{w}_j}:=(\theta(\mathbf{w}_j,\tilde{\mathbf{x}}_1),\dots, \theta(\mathbf{w}_j,\tilde{\mathbf{x}}_m))^T~.
    \end{equation}
    \item Given a quality level $\alpha$, select just the epochs for which the skill score $S$ computed on the validation set is bigger than $\alpha$. This allows the selection of the set of epochs
    \begin{equation}\label{eq:alpha}
        \mathcal{J}^*:=\{j\in\{1,\dots,N\} :  S(F_{\mathbf{\tilde{y}}}(I_{\tau_j^*}(\hat{\mathbf{y}}_{\tilde{\mathbf{X}}}^{\mathbf{w}_j}))) > \alpha\}.
    \end{equation}
    \item Define the result of the ensemble learning process as the binary value corresponding to the median value $m$ among all binary predictions associated to $\mathcal{J}^*$, i.e. given a new sample $\mathbf{x}$ the output is defined as
    \begin{equation}
        \hat{p}=m(\{I_{\tau_j^*}(\theta(\mathbf{w}_j,\mathbf{x})): j\in\mathcal{J}^*\}).
    \end{equation}
 In the case where the number of zeros is equal to the number of ones, we assume $\hat{p}=1$. 
\end{enumerate}

We now show how this process works in the case of three applications concerning pollution, space weather and stock market forecasting. Our focus will be on the assessment of results when a value-weighted skill score is used in equations (\ref{opt tau at j-th epoch}) and (\ref{eq:alpha}).

\subsection{Pollution forecasting}
\begin{figure}
    \centering
    \includegraphics[width=0.4\textwidth]{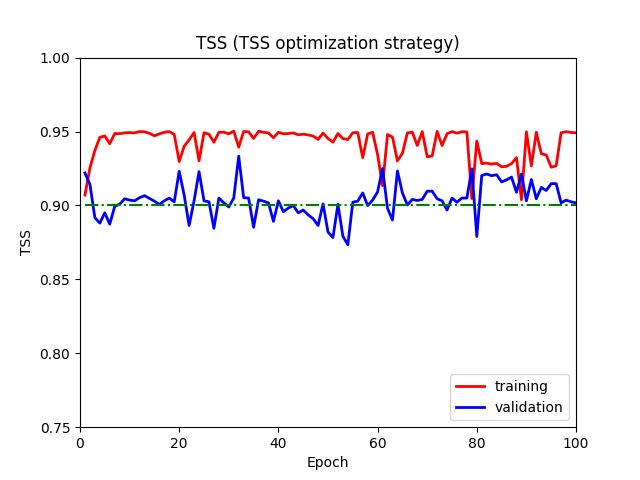}
    \includegraphics[width=0.4\textwidth]{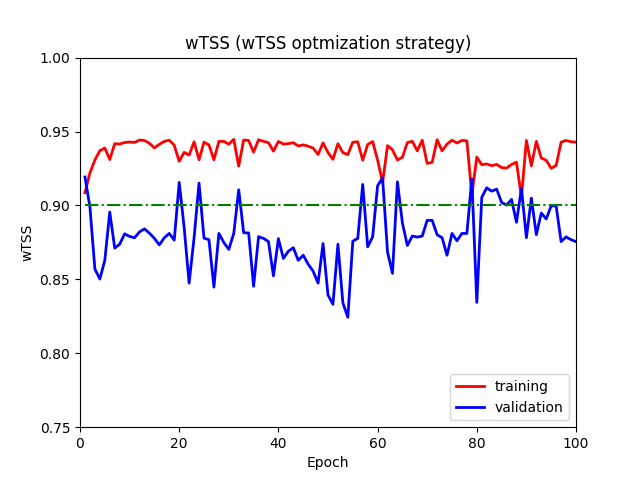}
    \caption{TSS and wTSS versus epochs, where the scores are computed on the training set (red line) and on the validation set (blue line) in the case of the PM5.2 pollution experiment. Left panel: TSS with optimization strategy involving TSS. Right panel: wTSS with optimization strategy involving wTSS.}
    \label{fig:ensemble_procedure}
\end{figure}

\begin{figure}
    \centering
    \subfigure[{TSS optimization strategy}]{
    \includegraphics[width=0.49\textwidth]{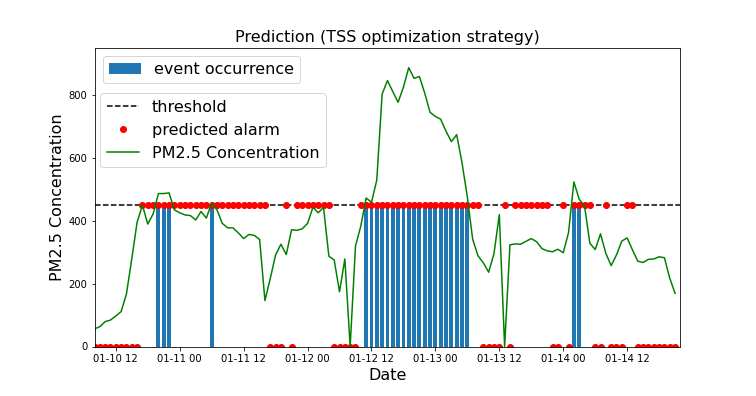}}
   \subfigure[{wTSS optimization strategy}]{\includegraphics[width=0.49\textwidth]{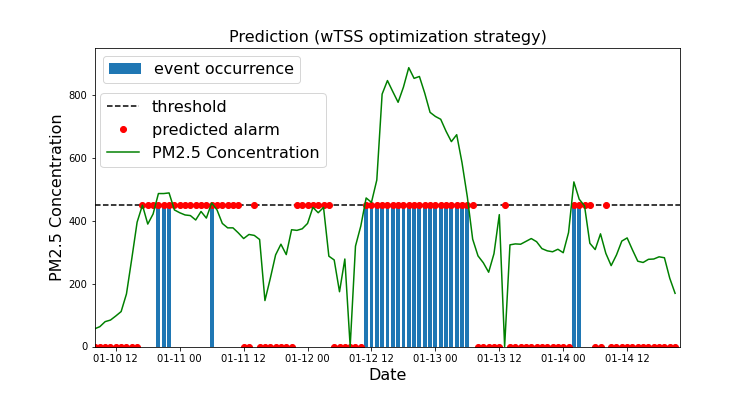}}\\
    \caption{PM2.5 pollution forecasting. Predictions provided by the ensemble method with the TSS optimization strategy (left panel) and the wTSS optimization strategy (right panel) on the period test from 1/10/2013 08:00 to 1/14/2013 21:00.}
    \label{fig:poll_ensemble_forecast}
\end{figure}

We consider the same data and the same forecasting problem discussed in Section \ref{sec:IV}, i.e. the prediction of over-threshold occurrences of PM2.5 concentration at time $T+1$, having at disposal measures of this concentration and of eight features associated to weather conditions at time samples from $0$ through $T$. The ensemble learning procedure is applied on the same training and validation sets as in Section \ref{sec:IV}. The computational core is now represented by a Long Short Term Memory (LSTM) neural network available in the Keras library \cite{chollet2015keras}, in which the sigmoid function and the binary cross-entropy are used as activation function and loss function, respectively. The model is trained over $N=100$ epochs using the Adam Optimizer \cite{Kingma2015AdamAM} with learning rate equal to $0.001$ and mini-batch size equal to $72$, respectively. As explained at the beginning of this section, the ensemble output is determined by combining estimators associated to epochs where the skill score computed on the validation set is higher than the quality level $\alpha$ (see equation \eqref{eq:alpha}). In this experiment $\alpha$ is fixed equal to $0.9$. Figure \ref{fig:ensemble_procedure} provides an illustrative example of the fact that using value-weighted skill scores leads to a stricter epoch selection strategy. Table \ref{tab:pollution_test} contains the results of the application of the ensemble learning approach on the test period from 12/29/2011 00:00 to 12/31/2014 23:00. Specifically, the table compares the classification performances provided by the optimization of wTSS versus the ones provided by the optimization of TSS. These numbers clearly show that optimizing the value-weighted skill score systematically leads to (sometimes even significantly) higher scores due to the fact that the number of wFPs is smaller than the number of FPs (while the numbers of wFNs and FNs do not change). Figure \ref{fig:poll_ensemble_forecast} offers a visualization enrolled over time of the behavior of the two optimization strategies on the test period considered in Section \ref{sec:IV}.

\begin{table}[ht]
		\centering
		\caption{PM2.5 pollution forecasting. Results on the test  period from 12/29/2011 00:00 to 12/31/2014 23:00 obtained by choosing the threshold for the ensemble method with the TSS optimization strategy and the wTSS optimization strategy. 
}
\label{tab:pollution_test}
\resizebox{0.5\textwidth}{!}{
\begin{tabular}{l l l l l }
& \multicolumn{4}{c}{Strategy}\\
\cline{2-5}
 & \multicolumn{2}{c}{Optimization of TSS} &  \multicolumn{2}{c}{ Optimization of wTSS}\\ 
Confusion matrix
& TP = 143 & FP = 785 & TP = 143  & FP = 400 \\ 
 & FN = 5  & TN = 25442 & FN = 5  & TN = 25827 \\ \hline 
TSS & \multicolumn{2}{c}{0.9363} &  \multicolumn{2}{c}{0.9510} \\ 
HSS & \multicolumn{2}{c}{0.2586} &  \multicolumn{2}{c}{0.4087}\\
CSI & \multicolumn{2}{c}{0.1533} & \multicolumn{2}{c}{0.2609}\\
wFN & \multicolumn{2}{c}{4.5} & \multicolumn{2}{c}{4.5} \\
wFP & \multicolumn{2}{c}{1401.42} & \multicolumn{2}{c}{650.42} \\
wTSS & \multicolumn{2}{c}{0.9173} & \multicolumn{2}{c}{0.9449} \\
wHSS & \multicolumn{2}{c}{0.1607} & \multicolumn{2}{c}{0.2974}\\
wCSI & \multicolumn{2}{c}{0.0923} & \multicolumn{2}{c}{0.1792}\\
\hline
\end{tabular}
}

\end{table}

\subsection{Solar flare forecasting}
Solar flares are the main trigger of space weather events, including Coronal Mass Ejections and solar wind \cite{tandberg1988physics}. Their prediction may rely on features extracted from magnetogram images of active regions (ARs) like the ones recorded by the {\em{Helioseismic and Magnetic Imager (HMI)}} on-board the {\em{Solar Dynamics Observatory (SDO)}} \cite{scherrer2012helioseismic}. We considered images from the {\em{HMI}} archives in order to setup an experiment in conditions analogous to the ones considered in 
\cite{Benvenuto2020MachineLearning,campi2019feature}:
\begin{itemize}
    \item We first grouped the images according to their issuing times and selected, in particular, just the images recorded at issuing time 00:00 UT.
    \item $23$ features were extracted by means of the algorithms illustrated in \cite{guerra2018active}. 
    \item The labelling process utilized the flare occurrence alarm sounded by the Geostationary Operational Environment Satellites (GOES) cluster and associated label $1$ to each feature vector where a flare was recorded within $24$ hours after the issuing time. GOES also computes the flare energetic class and, in particular, in the experiment we considered events of GOES class C1 and above (C1+ flares), M1 and above (M1+ flares) and X1 and above (X1+ flares). 
\end{itemize}
The training process exploited the {\em{HMI}} archive in the time range from 
09/15/2012 through 10/02/2015 and the validation process from 09/29/2015 through 01/11/2015 for the prediction of C1+ and M1+ solar flares. For the prediction of X1+ solar flares we considered a different splitting in order to have a reasonable number of positive samples in training and validation: we trained on the period from 09/15/2012 through 10/09/2014 and we validated on the period from 10/10/2014 through 06/14/2015.
For all three cases, the test phase focused on the time range from 01/13/2017 through 09/07/2017. We point out that, in this way, the validation set contains just two X1+ events, both associated to AR 12673 \cite{Benvenuto2020MachineLearning,guastavino2019desaturating}. In order to implement the ensemble learning approach we used a deep multi-layer perceptron with $7$ hidden layers. The Rectified Linear Unit (ReLU) function was used to activate the hidden layers, the sigmoid activation function was applied to activate the output and the binary cross-entropy was used as loss function. The model was trained over $100$ epochs using the Adam optimizer with learning rate equal to $0.001$, with a mini-batch size of $72$. In order to prevent overfitting, an $L^2$ regularization constraint was set as $0.01$ in the first two layers. The quality level $\alpha$ in equation (\ref{eq:alpha}) was fixed equal to a percentage of the maximum value of the skill score in the validation step: this rate was set equal to $90\%$ for prediction of C1+ flares, $95\%$ for prediction of M1+ flares and $80\%$ for prediction of X1+ flares. 

\begin{figure*}
    \centering
    \includegraphics[width=0.99\textwidth]{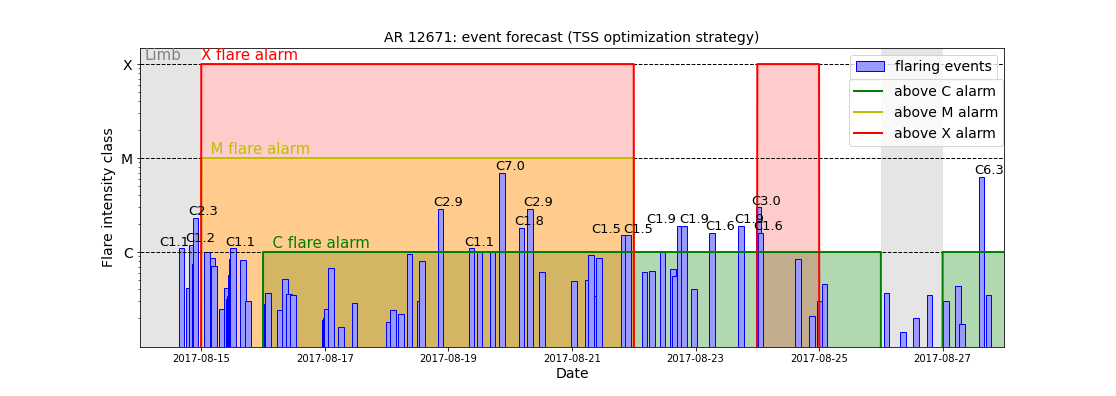}\\
    \includegraphics[width=0.99\textwidth]{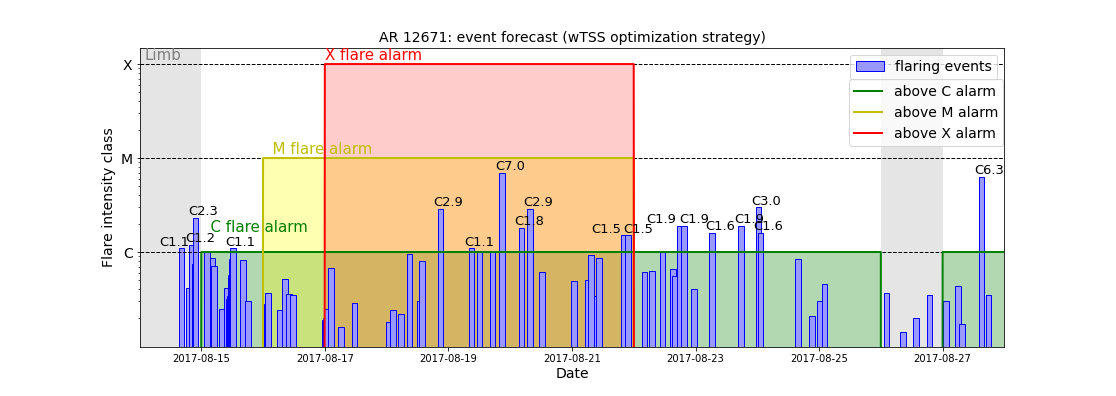}
    \caption{Predictions enrolled along time by the ensemble method when the TSS and wTSS optimization strategies are adopted (top and bottom panel, respectively). Green alarms correspond to C1+ flares, yellow alarms to M1+ flares, and red alarms to X1+ flares. The blue bar plots correspond to the actual flaring events recorded by GOES, the y-label representing the corresponding GOES flare
classes. Note that the grey boxes correspond to time period where the input data are missing.}
    \label{fig:ar12671}
\end{figure*}

The results of this analysis are shown in Table \ref{tab:solar_flare_test} and imply once again that the classification strategy based on the maximization of wTSS leads to higher scores (and more diagonal confusion matrices) with respect to the results provided by the maximization of TSS. This is particularly significant in the case of the prediction of M1+ and X1+ flares, i.e. in the case when the training set is significantly more imbalanced. Figure \ref{fig:ar12671} enrolls over time the prediction associated to AR 12671 \cite{boe2020cme,duncan2021nustar}, which originated many C1+ flares but no M1+ and no X1+ events. The figure clearly shows that the use of a value-weighted strategy leads to a significantly smaller number of FPs in the prediction of both M1+ and X1+ events.

\begin{table*}[ht]
		\centering
		\caption{Solar flare forecasting. Results on the test set obtained by realizing classification via TSS optimization (second, fourth and sixth column) and via wTSS optimization (third, fifth and seventh column). 
}\label{tab:solar_flare_test}
\resizebox{0.99\textwidth}{!}{
\begin{tabular}{l | l l l l l l l l l l l l}
& \multicolumn{4}{c|}{Prediction C1+ flares} & \multicolumn{4}{c|}{Prediction M1+ flares} & \multicolumn{4}{c}{Prediction X1+ flares}\\
\cline{2-5} \cline{6-9} \cline{10-13}
 & \multicolumn{2}{c}{Optimization of TSS} &  \multicolumn{2}{c|}{ Optimization of wTSS} & \multicolumn{2}{c}{Optimization of TSS} &  \multicolumn{2}{c|}{ Optimization of wTSS} & \multicolumn{2}{c}{Optimization of TSS} &  \multicolumn{2}{c}{ Optimization of wTSS}\\ 
Confusion matrix
& TP = 31 & FP = 30 & TP = 32  & \multicolumn{1}{c|}{FP = 32} & TP = 5 & FP = 21 & TP = 5  & \multicolumn{1}{c|}{FP = 19} & TP = 2 & FP = 31 & TP = 2  & FP = 20\\ 
 & FN =  3 & TN = 198 & FN = 2  & \multicolumn{1}{c|}{TN = 196} & FN =  1 & TN = 235 & FN = 1  & \multicolumn{1}{c|}{TN = 237} & FN =  0 & TN = 229 & FN = 0  & TN = 240\\ \hline 
TSS & \multicolumn{2}{c}{0.7802} & \multicolumn{2}{c|}{0.8008} & \multicolumn{2}{c}{0.7513} & \multicolumn{2}{c|}{0.7591} & \multicolumn{2}{c}{0.8808} & \multicolumn{2}{c}{0.9231}\\ 
HSS & \multicolumn{2}{c}{0.5832} & \multicolumn{2}{c|}{0.5823} & \multicolumn{2}{c}{0.2859} & \multicolumn{2}{c|}{0.3080}  & \multicolumn{2}{c}{0.1013} & \multicolumn{2}{c}{0.1548}  \\
CSI & \multicolumn{2}{c}{0.4844} & \multicolumn{2}{c|}{0.4848} & \multicolumn{2}{c}{0.1852} & \multicolumn{2}{c|}{0.2} & \multicolumn{2}{c}{0.0606} & \multicolumn{2}{c}{0.0909}\\
wFN & \multicolumn{2}{c}{2.5} & \multicolumn{2}{c|}{1.5} & \multicolumn{2}{c}{1} & 
\multicolumn{2}{c|}{1} & 
\multicolumn{2}{c}{0} & 
\multicolumn{2}{c}{0} \\
wFP & \multicolumn{2}{c}{29.92} & \multicolumn{2}{c|}{33.92} & \multicolumn{2}{c}{36.25} & \multicolumn{2}{c|}{32.25} & \multicolumn{2}{c}{60.5} & \multicolumn{2}{c}{40} \\
wTSS & \multicolumn{2}{c}{0.7941} & \multicolumn{2}{c|}{0.8077} & \multicolumn{2}{c}{0.6997} & \multicolumn{2}{c|}{0.7136} & \multicolumn{2}{c}{0.7910} & \multicolumn{2}{c}{0.8571}\\
wHSS &  \multicolumn{2}{c}{0.5886} & \multicolumn{2}{c|}{0.5715} &  \multicolumn{2}{c}{0.1807} & \multicolumn{2}{c|}{0.2012} &  \multicolumn{2}{c}{0.0494} & \multicolumn{2}{c}{0.0784} \\
wCSI & \multicolumn{2}{c}{0.4888} & \multicolumn{2}{c|}{0.4747} & \multicolumn{2}{c}{0.1183} & \multicolumn{2}{c|}{0.1307} & \multicolumn{2}{c}{0.032} & \multicolumn{2}{c}{0.0476}\\
\hline
\end{tabular}
}
\end{table*}

\subsection{Stock prize forecasting}
We considered the problem of predicting ``down" movements in stock prizes relying on information concerned with the daily closure prizes. More specifically, the feature utilized as input of the forecasting algorithm is a time series of five days of {\em{daily percentage change}} defined as \cite{chatfield2019analysis}
\begin{equation}\label{daily-percentage-change}
\eta = \frac{P_N - P_{N-1}}{P_{N-1}} \cdot 100~,
\end{equation}
where $P_{N-1}$ is the closure prize at day $N-1$ and $P_N$ is the closure prize at day $N$. We used as label for this feature the condition
\begin{equation}\label{condition}
\eta < L~,
\end{equation}
where $L=-1$ corresponds to the ``down" movement. We trained an LSTM NN on the training set in the time range from 10/01/2001 through 11/26/2007 in the database put at disposal by Yahoo Finance; the validation set is made of the same data, but in the time range from 11/27/2007 through 11/24/2009; the test set includes data from 11/25/2009 through 12/31/2010. 
We report in Table \ref{tab:finance_test} confusion matrices and skill scores corresponding to both the quality- and value-weighted approaches. These numbers show that, when we choose the wTSS optimization strategy, the ensemble learning method leads to predictions with lower TSS but higher wTSS. We further point out that, in stock index forecasting applications, the accuracy is often used for performance evaluation. Therefore, in table \ref{tab:finance_test} we also report both the quality- and value-weighted accuracy, although in our experiments the data sets are imbalanced, so that accuracy-type scores are less reliable than other skill scores like, for example, the TSS-type ones. Both accuracy and weighted value accuracy are slightly better for the strategy based on wTSS optimization.

\begin{figure*}
    \centering
    \includegraphics[width=0.99\textwidth]{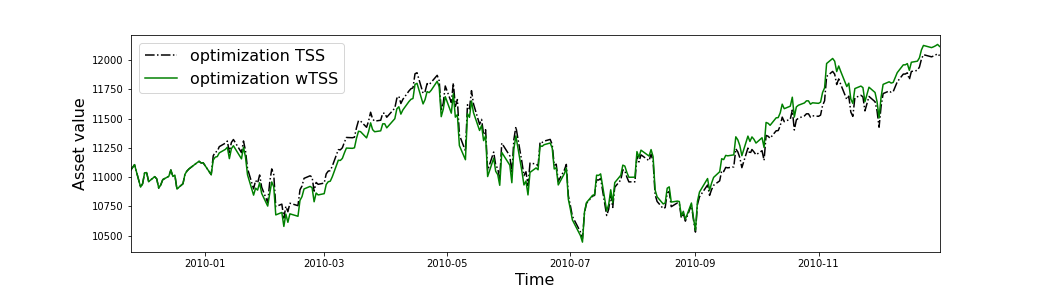}
    \caption{Asset profile versus time, associated to the quality-based and value-weighted optimization strategies, starting from an initial asset made of ten stocks. Dashed black line: the impact of the TSS optimization strategy. Green solid line: the impact of the wTSS optimization strategy.}
    \label{fig:finance}
\end{figure*}

In order to assess the value-weighted approach in an operational framework, we simulated the following investment strategy, starting from an initial asset of $10$ stocks:
\begin{enumerate}
    \item If at day $N-1$ a ``down" movement is predicted for day $N$, then we sell two stocks.
    \item If either at day $N$, or day $N+1$, or day $N+2$ the ``down" movement occurs, we use all the money earned at step 1 to buy stocks. At day $N+3$ we buy in any case. 
\end{enumerate}
This strategy is applied on the test set and the results of this analysis, illustrated in Figure \ref{fig:finance}, show that, in a long term perspective, the asset value provided by the value-weighted strategy overtakes the one provided by a standard quality-based strategy.

\begin{table}[ht]
		\centering
		\caption{Stock prize forecasting. Results on the test set extracted from the Yahoo Finance database, obtained by using the TSS and wTSS optimization strategies. 
}
\label{tab:finance_test}
\resizebox{0.5\textwidth}{!}{
\begin{tabular}{l l l l l }
& \multicolumn{4}{c}{Strategy}\\
 & \multicolumn{2}{c}{Optimization of TSS} &  \multicolumn{2}{c}{ Optimization of wTSS}\\ 
Confusion matrix
& TP = 22 & FP = 78 & TP = 20  & FP = 70 \\

 & FN = 19  & TN = 158 & FN = 21  & TN = 166 \\ \hline 
TSS & \multicolumn{2}{c}{0.2061} &  \multicolumn{2}{c}{0.1912} \\ 
ACC & \multicolumn{2}{c}{0.6498} &  \multicolumn{2}{c}{0.6715}\\
CSI & \multicolumn{2}{c}{0.1849} & \multicolumn{2}{c}{0.1802}\\
wFN & \multicolumn{2}{c}{17} & \multicolumn{2}{c}{17.42} \\
wFP & \multicolumn{2}{c}{71.83} & \multicolumn{2}{c}{62.33} \\
wTSS & \multicolumn{2}{c}{0.2516} & \multicolumn{2}{c}{0.2615} \\
wACC & \multicolumn{2}{c}{0.67} & \multicolumn{2}{c}{0.7}\\
wCSI & \multicolumn{2}{c}{0.1985} & \multicolumn{2}{c}{0.2005}\\
\hline
\end{tabular}
}
\end{table}

\section{Comments and conclusions}
\label{sec:VI}
This study introduces two novelties in the machine learning game. 

The first one is a way to assess the forecasting performances of a machine learning algorithm, in which false positives anticipating the actual event occurrence are weighed less than the ones associated to alarms sounded behind schedule. This approach allows the definition of value-weighted skill scores that evaluate the forecasting performances in a way which is more appropriate for decision making processes. 

The second novelty is related to the optimization of ensemble learning for classification purposes. Indeed, in three different forecasting contexts, we show that when the probabilistic outcomes of a NN are clustered by optimizing a value-weighted skill score, then the forecasting performances are systematically more reliable than the ones provided via optimization of a standard quality-based skill score. 

The next step in this investigation will be the encoding of these value-weighted newly introduced skill scores in the loss functions utilized as part of the ensemble learning algorithm, in such a way to implement a forecasting approach in which the value-weighted strategy is {\it a priori} introduced in the optimization process.



%

\appendices


\section*{Acknowledgment}
SG is financially supported by a regional grant of the `Fondo Sociale Europeo', Regione Liguria. 
MP and FB acknowledge the financial contribution from the agreement ASI-INAF n.2018-16-HH.0.

\ifCLASSOPTIONcaptionsoff
  \newpage
\fi



\bibliography{reference.bib}
\bibliographystyle{IEEEtran}
\end{document}